\documentclass{article}
\usepackage{spconf,amsmath,graphicx,hyperref}
\usepackage{multirow}
\usepackage{xcolor}
\usepackage{amssymb}
\usepackage[english]{babel}
\usepackage{booktabs,xcolor,colortbl,array}
\usepackage{fancyhdr}

\newcommand{\refcell}[1]{\cellcolor{gray!20}{#1}}
\definecolor{baseline}{RGB}{240,240,240}
\definecolor{pom}{RGB}{232,241,255}
\definecolor{comp}{RGB}{255,243,230}
\definecolor{lmcol}{RGB}{248,248,248}
\newcolumntype{L}{>{\columncolor{lmcol}}c}
\newcolumntype{C}[1]{>{\centering\arraybackslash}m{#1}}     
\newcolumntype{R}[1]{>{\centering\arraybackslash}m{#1}}
\newcommand{\rowlabel}[1]{\rotatebox[origin=c]{90}{\normalsize #1}}

\newcommand{\basecell}[1]{\cellcolor{baseline}{#1}}
\newcommand{\pomcell}[1]{\cellcolor{pom}{#1}}
\newcommand{\compcell}[1]{\cellcolor{comp}{#1}}

\title{Polynomial mixing for efficient self-supervised speech encoders}
\name{Eva Feillet$^{1,2}$\thanks{This work was performed using HPC resources from GENCI–IDRIS (Grant 2025-AD011016345 and 2024-AD011014732). EF received funding from the French ANR project N°ANR-23-PEIA-0008 (SHARP) within the France 2030 program. RW received funding from the French ANR project N°ANR-22-CE23-0013 (E-SSL).}, Ryan Whetten$^3$, David Picard$^4$, Alexandre Allauzen$^2$}
\address{$^1$Université Paris-Saclay, CNRS, Laboratoire Interdisciplinaire des Sciences du Numérique \\
$^2$Miles team, Université Paris-Dauphine-PSL 
$\;^3$Laboratoire Informatique d'Avignon,\\ Avignon Université 
$\;^4$ LIGM, École Nationale des Ponts et Chaussées
}
\begin{document}
\ninept
\maketitle
\pagestyle{fancy}
\fancyhf{} % clear header/footer
\fancyfoot[L]{\footnotesize \textit{\textcopyright 2026 IEEE. Personal use of this material is permitted. Permission from IEEE must be obtained for all other uses, in any current or future media, including reprinting/republishing this material for advertising or promotional purposes, creating new collective works, for resale or redistribution to servers or lists, or reuse of any copyrighted component of this work in other works.}}
\renewcommand{\headrulewidth}{0pt} % optional: no header rule

\thispagestyle{fancy} % important: \maketitle often forces page 1 to 'empty'

\begin{abstract}
State-of-the-art speech-to-text models typically employ Transformer-based encoders that model token dependencies via self-attention mechanisms. 
However, the quadratic complexity of self-attention in both memory and computation imposes significant constraints on scalability. 
In this work, we propose a novel token-mixing mechanism, the \textbf{Polynomial Mixer} (PoM), as a drop-in replacement for multi-head self-attention. 
PoM computes a polynomial representation of the input with linear complexity with respect to the input sequence length.
We integrate PoM into a self-supervised speech representation learning framework based on BEST-RQ and evaluate its performance on downstream speech recognition tasks. Experimental results demonstrate that PoM achieves a competitive word error rate compared to full self-attention and other linear-complexity alternatives, offering an improved trade-off between performance and efficiency in time and memory. %(\textcolor{blue}{use a composite score?})
\end{abstract}

% keywords can be removed

\section{Introduction}
\label{sec:intro}

Transformer-based architectures have become a cornerstone of modern speech recognition, with leading self-supervised methods such as wav2vec 2.0 \cite{baevski2020wav2vec}, BEST-RQ \cite{chiu2022bestrq}, on which for example Google USM~\cite{zhang2023google} is based, and the Whisper family of models~\cite{radford2023robust}. 

Although some of these methods take advantage of specialized variants of the original Transformer \cite{vaswani2017attention} (e.g., Conformer \cite{gulati2020conformer}), they all rely on multi-head attention (MHA), whose quadratic cost in the input sequence length still remains a major computational bottleneck~\cite{whetten2024analysis}.

Numerous alternative \textit{token mixers} have been proposed, e.g., MLP-Mixer \cite{tolstikhin2021mlp} in computer vision, BigBird \cite{zaheer2020big}, Linformer \cite{wang2020linformer}, and FastFormer \cite{wu2021fastformer} in natural language processing. State-space models, such as Mamba \cite{gu2023mamba}, also offer an efficient alternative to MHA. 
In contrast, speech recognition has received comparatively little attention in this direction. To date, few works have proposed speech-specific alternatives, for instance SummaryMixing \cite{parcollet2024summarymixing}. 
In this article, we introduce the Polynomial Mixer (PoM), a novel speech-tailored token mixer with linear complexity in the number of input tokens. Building on the insight of SummaryMixing that accurate speech recognition is possible without computing pairwise token interactions exhaustively,
PoM is designed as a drop-in replacement for attention while retaining sufficient expressivity for the complexity of spoken language.

We pre-train various encoders on LibriSpeech-960h~\cite{panayotov2015librispeech} following BEST-RQ self-supervised learning scheme and show that PoM comes close to full attention mechanisms. 
PoM also shows competitive runtime and memory use compared to linear alternatives and surpasses SummaryMixing in word error rate when fine-tuned on LibriSpeech-100h. 
We will release the code as a plug-in to SpeechBrain Toolkit~\cite{ravanelli2021speechbrain}.

\section{Related work}
\label{sec:related}

\subsection{Self-supervised speech recognition models}
Self-supervised learning (SSL) has become the dominant paradigm for pre-training speech encoders, taking advantage of large-scale data without transcription. 
For example, wav2vec 2.0 \cite{baevski2020wav2vec} introduced SSL from raw audio data through a contrastive task and demonstrated the feasibility of performing automatic speech recognition (ASR) from a few labeled data.
In the same vein, WavLM's \cite{chen2022wavlm} pretraining task consists of jointly predicting masked speech segments and denoising in order to improve the transferability of the learnt representation to a wider range of downstream tasks.  
More recently, BEST-RQ \cite{chiu2022bestrq} proposed an efficient alternative to previous SSL approaches by using Mel filter banks as inputs rather than raw audio data. Therefore, it does not have to learn extra convolutional layers to encode the input.
BEST-RQ relies on a fixed, randomly initialized codebook to assign target labels to input frames. 
Its self-supervised training process consists of training the encoder to predict, for a given input frame, the index of the vector in the codebook that is closest to the frame.
In this article, we chose BEST-RQ as a pre-training method because it has been shown to achieve state-of-the-art performance with improved efficiency compared to previous SSL schemes, such as wav2vec 2.0. Additionally, BEST-RQ has an open source implementation in the SpeechBrain toolkit \cite{whetten2024implem}. 

From an architectural perspective, most state-of-the-art speech encoders adopt Transformer-based designs. For example, this is the case of the BEST-RQ original implementation, which is based on the Conformer architecture \cite{gulati2020conformer}, which combines convolutional and self-attention layers sequentially to better capture local and global dependencies. %Similarly, the Branchformer \cite{peng2022branchformer} introduces parallel branches of attention and MLP layers to improve modeling efficiency. 
These encoders remain reliant on MHA and thus inherit its quadratic complexity. This reliance motivates the development of more efficient alternatives to MHA in speech representation learning.
In this paper, we propose to replace the default attention mechanism of the Conformer blocks by a \emph{polynomial token mixer}.

\subsection{Token mixing methods}
Currently, the dominant approach to token mixing in sequence modeling is MHA \cite{vaswani2017attention}, which computes pairwise interactions across all tokens. Several extensions have been proposed to increase the context range of MHA or enhance its positional encodings. For example, \cite{dai2019transformer} proposes relative positional encoding (RelPos) for handling longer dependencies and Rotary Position Embeddings (RoPE) \cite{su2024roformer} encodes relative positions using rotation matrices.

Significant efforts have been made towards reducing the computational cost of MHA. Sparse attention mechanisms, such as BigBird \cite{zaheer2020big} and Longformer \cite{beltagy2020longformer}, reduce computational cost by computing only a subset of the attention coefficients. Low-rank approximations, such as Linformer \cite{wang2020linformer}, compress attention maps into lower-dimensional projections, but lose the ability to process sequences of varying length in the process. The Performer \cite{choromanski2021performer} and kernel-based linear attention methods \cite{katharopoulos2020lin} achieve linear complexity by reinterpreting attention through kernel functions. Other strategies approximate global interactions via summary vectors, as in FastFormer \cite{wu2021fastformer}.
Beyond attention-inspired methods, alternative token mixers include MLP-based approaches such as MLP-Mixer \cite{tolstikhin2021mlp} and HyperMixer \cite{mai2023hypermixer}, which replace attention with feed-forward mixing operations. Non-parametric approaches have also been proposed, most notably FNet \cite{lee2022fnet}, which leverages Fourier transforms to mix tokens. 

In speech, token mixing poses specific challenges compared to text. Speech recognition models must handle long input sequences, comprising tokens corresponding to milliseconds of audio, while producing relatively short output sequences of letters or phonemes. Besides, as noted by Zhang et al. \cite{zhang2021usefulness}, modeling long-range dependencies may be less critical for ASR, since local acoustic information often suffices for predicting individual output units. This reduces the necessity of global attention across all frames.

The works closest to ours are SummaryMixing \cite{parcollet2024summarymixing}, which achieves linear complexity by summarizing the input sequence into an average vector, which is concatenated back to the input tokens, and the recent analysis of linear-complexity alternatives to attention for speech by Whetten et al. \cite{whetten2024analysis}. 
Finally, on the side of computer vision, \cite{picard2024pom} introduces a new family of token-mixing mechanisms that, like SummaryMixing, departs from explicit pairwise interactions and inspires the present work.
This motivates us to present our polynomial mixing method.

\section{Method}
\label{sec:method}
\begin{figure}
    \centering
    \includegraphics[width=0.95\linewidth]{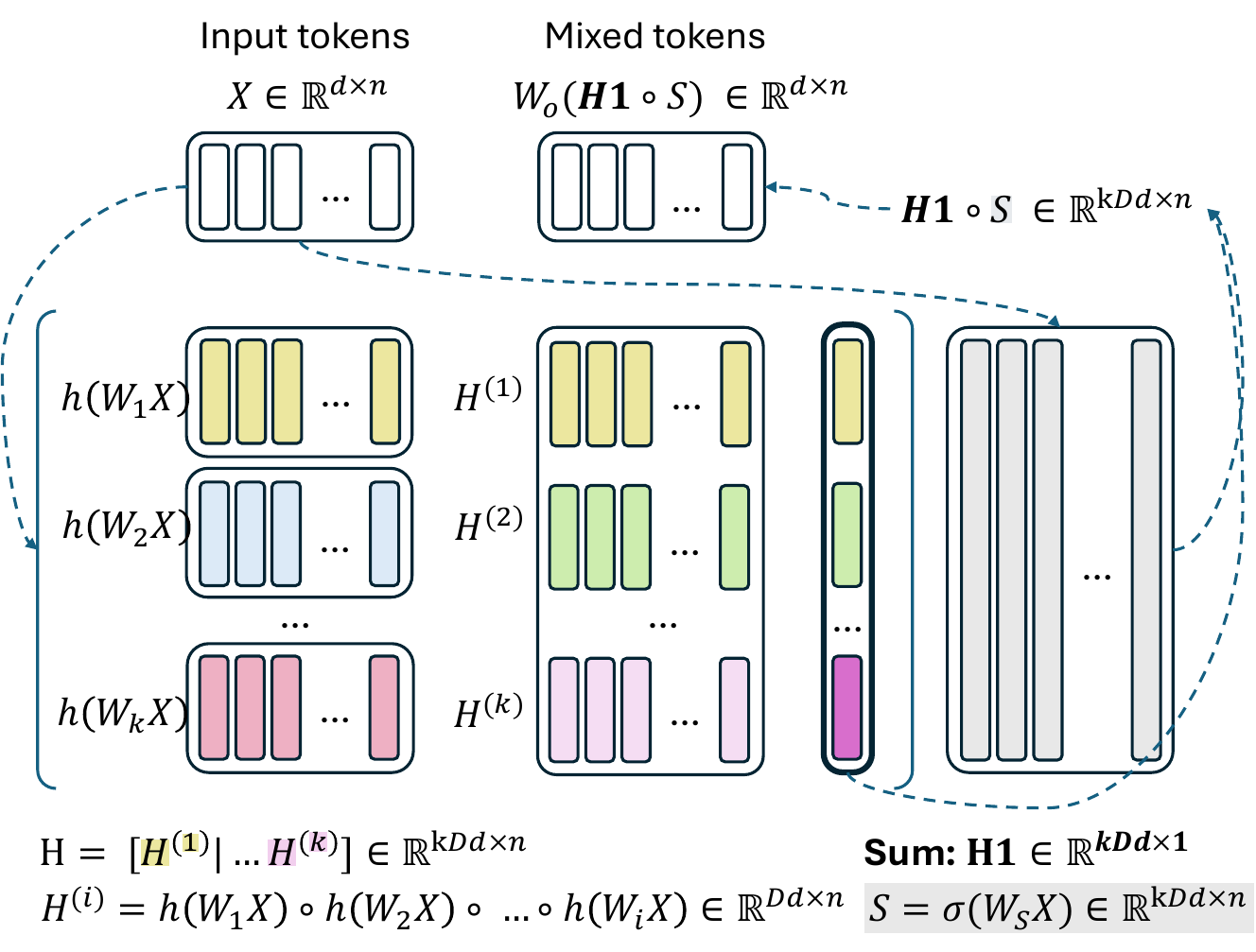}
    \caption{Principle of the Polynomial Mixer. The input sequence is projected through $k$ polynomial branches, aggregated into a global representation $H(X)$, and combined with a token-wise selector $S$. The output is obtained by projecting the selected state back to the input space.} 
    \label{fig:principle}
\end{figure}

%\subsection{Self-supervised learning with BEST-RQ}
\subsection{Polynomial Mixer for speech}
Taking inspiration from recent work in vision, we propose to use the \textit{Polynomial Mixer} (PoM), a sequence-to-sequence operator with linear complexity designed to replace MHA. 

\textbf{Definition.} As depicted in Figure~\ref{fig:principle}, the PoM maps an input matrix $X \in \mathbb{R}^{d \times n}$ to an output matrix in $\mathbb{R}^{d \times n}$ by summarizing the sequence into a state representation and broadcasting relevant information back to each token.
The summary is obtained by computing a polynomial of degree $k$ in an embedding space that has a dimension expanded by a factor $D$. 
Given an input $X \in \mathbb{R}^{d \times n}$, representing a sequence of $n$ tokens of dimension $d$, the Polynomial Mixer is  defined as:
\begin{equation} \label{eq:PoM}
\text{PoM}(X) = W_o \left[ \sigma(W_s X) \circ H(X) \mathbf{1}^{\top} \right],
\end{equation}
where $\circ$ denotes the element-wise (Hadamard) product, $\sigma$ is an element-wise sigmoid activation, and $\mathbf{1} \in \mathbb{R}^{n \times 1}$ is a vector of ones. 
Using learnable projection matrices $W_s \in \mathbb{R}^{kD \times d}$ and $W_o \in \mathbb{R}^{d \times kD}$, the input is successively projected in a higher-dimensional space, where it is combined with a summary of the sequence, and projected back to its original dimension. \\
\textbf{Global representation $H(X)$.} The key idea of PoM is to compute a \textit{global state representation} $H(X) \in \mathbb{R}^{kD \times n}$ by mixing all tokens through a fixed-degree polynomial of their projections. The function $H(X)$ defines the state representation $H(X) \in \mathbb{R}^{D d k \times n}$ as follows:
{\footnotesize \begin{equation} \label{eq:H}
H(X) = \left[ h(W_1 X) \;\middle|\; h(W_1 X) \circ h(W_2 X) \;\middle|\; \dots \;\middle|\; \prod_{m=1}^k h(W_m X) \right]\mathbf{1}
\end{equation}}
where $W_1, \dots, W_k \in \mathbb{R}^{D \times d}$ are learnable parameters, $h(\cdot)$ is a non-linear activation function (GELU in our implementation), $|$ denotes vertical concatenation and the product is computed element-wise across the projected views. Processed tokens are then summed.\\
\textbf{Token-wise selector $S$.} Each input token then attends to the shared state using a learned query matrix $W_s$:
\begin{equation} \label{eq:S}
S = \sigma(W_s X) \in \mathbb{R}^{kD \times n}.
\end{equation}
The product $H(X) \mathbf{1}^{\top} \in \mathbb{R}^{kDd \times n}$ replicates the shared state across all time steps, and the element-wise product $S \circ (H(X) \mathbf{1}^{\top})$ selects relevant components of the state per token. The final output is projected back to the input dimension with $W_o$.

\textbf{A drop-in replacement for MHA.} To integrate PoM into an encoder, we design a block $P$ as an alternation of a PoM and a feed-forward layer, with residual connections:
\begin{equation} \label{eq:block}
P(X) = X + \text{PoM}(X) + \text{FF}(X + \text{PoM}(X)),
\end{equation}
where $\text{FF}(\cdot)$ is a standard two-layer feed-forward network. 
This block is a drop-in replacement for standard Transformer encoder blocks. 
Our experiments with a Conformer show that PoM retains the sequence-to-sequence mapping capacity of MHA.

\textbf{Complexity.}
Unlike MHA, which computes all pairwise interactions between tokens at quadratic cost in sequence length $n$, the PoM relies on a global state that all tokens access independently. This leads to a \textbf{linear} complexity in $n$, both in time and memory. 

\subsection{Variants of PoM} \label{subsec:pom_variants}
\textbf{Mode jump.} We experiment with a variant of PoM consisting of using only the highest degree $k$ instead of keeping all the degrees up to $k$. 
In the standard PoM, the polynomial feature map is defined as the concatenation of nonlinear projections up to order $k$ (Equation \ref{eq:H}). 
In this variant, we have
\[
H^{(k)} =  \left[\prod_{m=1}^k h(W_m X) \right]\mathbf{1} \in \mathbb{R}^{D d \times n},
\]
which simplifies the representation and saves parameters in projection matrices $W_S$ and $W_O$. 
\textbf{Selective PoM.} We further propose a variant of PoM that applies the polynomial operation selectively on only half of the input features. The motivation is to force the model to perform mixing only on the features relevant to this operation, while keeping local information unmixed. 

\textbf{Mixing high and low frequencies separately.}
We introduce a frequency-aware PoM, which splits the input along the feature dimension before applying mixing separately to each group of features within each PoM block. 
In this case, we split the input $X \in \mathbb{R}^{d \times n}$ along the feature dimension : $ X = [ X^{'} | X^{''}]$, with $d = d' + d''$, $X^{'} \in \mathbb{R}^{d' \times n}$ and $X^{''} \in \mathbb{R}^{d'' \times n}$.
Then for each polynomial branch $i$, instead of computing $h(W_i X)$, we compute:
$$
\left[h(W_i^{'} X^{'}) \;\middle|\; h(W_i^{''} X^{''})\right],
$$ %\sigma(W_i X) \longrightarrow
where
$W_i^{'} \in \mathbb{R}^{d' \times Dd'}$ and 
$W_i^{''} \in \mathbb{R}^{d'' \times Dd''}$.
Concatenation yields an output in $\mathbb{R}^{ D \times n}$. 
This modification is applied independently for each branch $i$, so that we obtain two global representations $H(X^{'})$ and $H(X^{''})$. Selection $S$ is also applied separately. Notice that all features can still interact through the skipped connection.  
We interpret this operation as providing separate mixing paths for high and low frequencies, encouraging the learning of different parameters for semantic vs. phonemic content.

\section{Results}\label{sec:experiments}

\subsection{Experimental setting}
We implement PoM in PyTorch as a plug-in for the Speechbrain library \cite{ravanelli2021speechbrain} (version 1.0.3)\footnote{Code available at \url{https://github.com/EvaJF/pom4speech}.} %. We release the code and configuration files for reproducing the experiments. 

\textbf{Baselines.} 
We compare PoM with several variants of MHA implemented in SpeechBrain, namely regular MHA and the more competitive relative positional (RelPos) encoding \cite{dai2019transformer} and rotary positional encoding (RoPE) \cite{su2024roformer}. 
SummaryMixing was shown by \cite{whetten2024analysis} to obtain one of the best trade-offs between runtime, memory and word error rate (WER) in ASR, so we reproduce results with this method, and report the results of \cite{whetten2024analysis} with other token mixers of linear complexity (HyperConformer, Performer and Mamba). We note that our reproduction of SummaryMixing yields slightly lower performance than \cite{whetten2024analysis} when the language model is omitted, and higher performance when it is included. Hypermixing is another competitive baseline, but we were unable to reproduce the HyperConformer results due to convergence failures. Note that we ended pretraining of the large models earlier than \cite{whetten2024analysis}, which may advantage the corresponding unreproduced entries in Table~\ref{tab:asr}.

\textbf{Encoder pre-training.}
We pre-train encoders using the BEST-RQ implementation by \cite{whetten2024implem}.
The basic architecture is a Conformer, and alternative token mixers (PoM and its variants, SummaryMixing) are directly integrated into the Conformer architecture as drop-in replacements for MHA. 
The encoders are pre-trained on the 960h-LibriSpeech dataset \cite{panayotov2015librispeech}, consisting of English audiobooks. 
The encoder of ``base'' models ($\sim95$M parameters) is composed of 12 blocks and are trained for 200k steps on 4 A100 GPUs with a batch size of $1400$s per GPU, accounting for a total batch size of $1.6$h. 
``Large'' models ($\sim315$M parameters) are composed of 24 encoder blocks and are trained for 200k steps with a total batch size of $6400$s or $1.8$h. Note that this number of training steps and batch size is relatively small compared to current standards to reach state-of-the-art WER on LibriSpeech. 
However, in accordance with \cite{whetten2025early}, we have found ASR rankings across mixer types to be stable across pre-training epochs once encoders have started to converge on the pre-training task. And as underlined by \cite{parcollet2021energy}, small improvements towards state-of-the-art results may be obtained at a very high compute cost. Hence, our choice of a relatively limited number of pre-training steps.
We used $k=3, D=1$ and $k=3, D=2$ for the 95 and 315 million-parameter PoM models, respectively.

%Downstream task
\textbf{ASR fine-tuning.} We compare the pre-trained encoders by applying them in combination with a decoder to the task of automatic speech recognition (ASR).
Encoders are fine-tuned on the LibriSpeech-100h ``clean'' subset for 30 epochs with a 3-layer linear decoder and CTC loss. We chose this simple decoder to better show the impact of the encoder on the downstream ASR task.
Models are evaluated on the \textit{test-clean} and \textit{test-other} subsets, with and without language model \cite{heafield2013kenLM}. %the ``kenLM'' 

%-------------------------------------------------%
\subsection{Main results}

\begin{figure}
        \centering
        \includegraphics[width=1\linewidth]{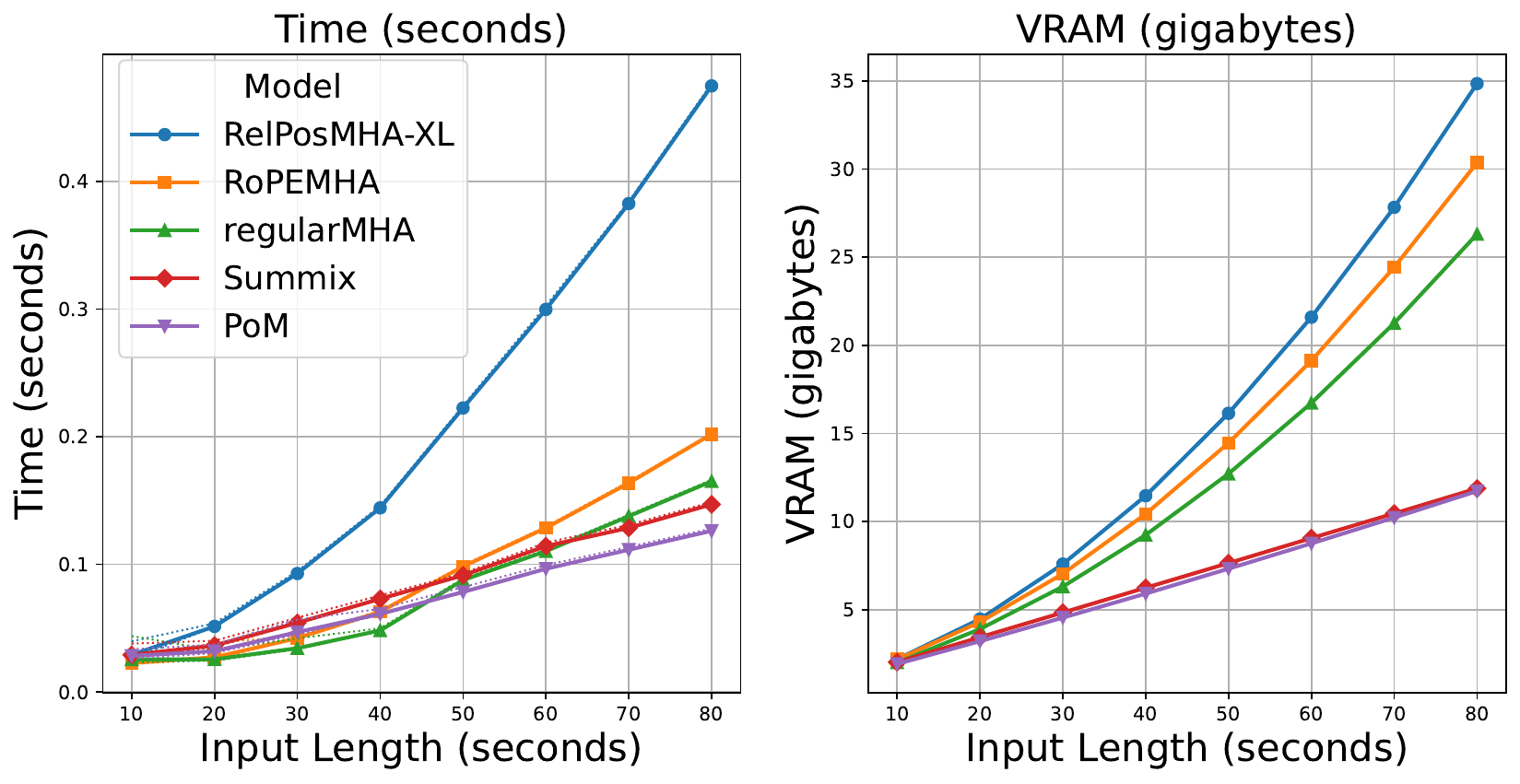}
        \caption{Inference time and peak memory usage of BEST-RQ models ($\sim95$M params) with various token mixers. Input length is
increased from 10 to 80 seconds. MHA requires significantly more time and VRAM as the input size
increases in comparison with linear alternatives, including PoM.}
        \label{fig:monitoring}
\end{figure}

As shown in Table \ref{tab:asr}, \textbf{PoM 95M-parameter model outperforms other linear alternatives and obtains competitive WER with regular MHA.} 
\textbf{PoM performance also scales with model size.} 
SummaryMixing is the linear mixer that is closest to PoM in terms of functioning. However, we argue that PoM is more expressive than SummaryMixing because the state that summarizes the input sequence in PoM benefits from higher-order interactions. In contrast, SummaryMixing only computes an arithmetical mean for its summary vector. 
Mamba and HyperConformer are strong baselines for replacing MHA with a linear-complexity mixer. PoM is competitive with them, as no method is best in all settings (test subset, language model). 
The WER obtained by RelPosMHA and RoPE is hard to match, however, as shown in Figure \ref{fig:monitoring}, it comes at a significantly higher cost in terms of runtime and memory.

\textbf{Without particular implementation optimization, PoM uses 2.8 times less memory than RelPosMHA for an $80s$-input sequence.} 
PoM has a runtime close to SummaryMixing and is also faster than RoPE, despite RoPE using the optimized PyTorch implementation of multi-head attention. We note that the SpeechBrain implementation of RelPosMHA does not enjoy the same optimization as RoPE and regularMHA.

\begin{table}[h]
\centering 
\small
\caption{WER on LibriSpeech \textit{test-clean} and \textit{test-other}. Models are pretrained on LibriSpeech-960h and fine-tuned on \textit{train-100}. Confidence intervals from 1000 bootstrap trials. Results with $^\dagger$ reported from \cite{whetten2024analysis}. Lower is better. \textbf{Best MHA variant} in bold, \underline{best linear mixer} underlined.}
\setlength{\tabcolsep}{4pt}
\resizebox{0.50\textwidth}{!}{
\begin{tabular}{R{3mm}|l c c c c}
\toprule
& \textbf{Model} & Clean & Clean + LM & Other & Other + LM \\
\midrule
\multirow{8}{*}{\rowlabel{$\sim$95M param.}}
 & \basecell{RelPosMHA}                & \basecell{\textbf{7.96 ($\pm$ 0.32)}} & \basecell{\textbf{4.89 ($\pm$ 0.25)}} & \basecell{\textbf{17.61} ($\pm$ 0.54)} & \basecell{12.13 ($\pm$ 0.44)} \\
 & \basecell{RoPE MHA}                 & \basecell{\textbf{8.06} ($\pm$ 0.31)} & \basecell{\textbf{4.90} ($\pm$ 0.26)} & \basecell{\textbf{17.53 ($\pm$ 0.48)}} & \basecell{\textbf{11.98 ($\pm$ 0.45)}} \\
 & \basecell{regular MHA}              & \basecell{8.59 ($\pm$ 0.32)} & \basecell{5.37 ($\pm$ 0.25)} & \basecell{19.44 ($\pm$ 0.54)} & \basecell{13.47 ($\pm$ 0.46)} \\
 & \pomcell{PoM base}              & \pomcell{8.31 ($\pm$ 0.31)} & \pomcell{\underline{5.42} ($\pm$ 0.27)} & \pomcell{\underline{19.06} ($\pm$ 0.53)} & \pomcell{\underline{13.62} ($\pm$ 0.48)} \\
 & \compcell{SummaryMixing}            & \compcell{9.79 ($\pm$ 0.34)} & \compcell{5.93 ($\pm$ 0.27)} & \compcell{22.80 ($\pm$ 0.60)} & \compcell{15.84 ($\pm$ 0.51)} \\
 & \compcell{Mamba$^\dagger$}          & \compcell{\underline{7.61}} & \compcell{\underline{5.50} ($\pm$ 0.28)} & \compcell{19.97} & \compcell{15.37} \\
 & \compcell{HyperConformer$^\dagger$} & \compcell{8.22} & \compcell{5.77 ($\pm$ 0.28)} & \compcell{19.29} & \compcell{15.03} \\
 & \compcell{FastFormer$^\dagger$}     & \compcell{9.32} & \compcell{6.82 ($\pm$ 0.31)} & \compcell{22.75} & \compcell{17.95} \\
\midrule
\multirow{7}{*}{\rowlabel{$\sim$315M param.}}
 & \basecell{RelPosMHA}                & \basecell{\textbf{4.92 ($\pm$ 0.25)}} & \basecell{\textbf{3.49 ($\pm$ 0.21)}} & \basecell{\textbf{10.78 ($\pm$ 0.37)}} & \basecell{\textbf{8.09 ($\pm$ 0.35)}} \\
 & \basecell{RoPE MHA}                 & \basecell{5.13 ($\pm$ 0.26)} & \basecell{3.66 ($\pm$ 0.21)} & \basecell{10.99 ($\pm$0.33)} & \basecell{8.45 ($\pm$ 0.37)}  \\ 
 & \pomcell{PoM base}              & \pomcell{6.28 ($\pm$ 0.26)} & \pomcell{\underline{4.52} ($\pm$ 0.23)} & \pomcell{14.86 ($\pm$ 0.47)} & \pomcell{\underline{11.33} ($\pm$ 0.43)} \\
 & \compcell{SummaryMixing}            & \compcell{7.35 ($\pm$ 0.31)} & \compcell{4.85 ($\pm$ 0.25)} & \compcell{17.60 ($\pm$ 0.53)} & \compcell{12.97 ($\pm$ 0.49)} \\
 & \compcell{Mamba$^\dagger$}          & \compcell{\underline{5.59}} & \compcell{\underline{4.48} ($\pm$ 0.25)} & \compcell{15.47} & \compcell{12.66} \\
 & \compcell{HyperConformer$^\dagger$} & \compcell{5.87} & \compcell{\underline{4.54} ($\pm$ 0.32)} & \compcell{\underline{13.13}} & \compcell{\underline{10.78}} \\
 & \compcell{FastFormer$^\dagger$}     & \compcell{13.16} & \compcell{9.89 ($\pm$ 0.34)} & \compcell{31.91} & \compcell{26.75} \\
\bottomrule
\end{tabular}
}
\label{tab:asr}
\end{table}

\subsection{Ablation study} 

\textbf{PoM components.} We train different configurations of PoM with the combinations of degree $k \in \{1, 2, 3\}$, expansion factor $D \in \{1, 2, 3\}$, hidden size $d \in \{488, 512, 576, 616 \}$, and versions of PoM presented in Subsection \ref{subsec:pom_variants}. The resulting models range from 75 to 122 million parameters. Their WER is averaged by PoM version in Table \ref{tab:ablation_pom}  
We find PoM ASR performance to increase with the product of $k$, $D$ and $d$ but to saturate around $k=2$ and $D=2$ for a fixed number of parameters (95M). As reported in Table \ref{tab:ablation_pom}, 
jumping modes (``select'') harm ASR performance. This underlines the interest of the polynomial mixing approach. 
Models trained with separated mixing of different groups of frequencies (splitting frequencies either into two or three groups / ``2ways'', ``3ways'') showed slight WER improvements for underfitting encoders but no improvement when continuing pre-training. Finally, we kept the ``base'' version of PoM.

\textbf{Choice of encoding.} In preliminary experiments, we did not observe significant improvement for PoM when applying RoPE encoding instead of fixed sinusoidal encoding. This is coherent with the fact that PoM is based on a Hadamard product, whereas RoPE multiplicative, relative, encoding is useful in the case of dot-product attention. 

\textbf{Use of layer drop.}
The results reported in Table \ref{tab:asr} are obtained by applying a $5\%$ layer drop during pre-training and fine-tuning, as we found that layer drop benefits all mixers. 
In Table \ref{tab:ablation_training}, we analyse the impact of layer drop on WER, in combination with the use of a language model.
All encoders have around $95$ million parameters. Regular MHA, RoPE, and RelPosMHA are grouped under ``MHA''. The various configurations of PoM ``base'' (varying $k$ and $D$) are aggregated under ``PoM''.
We observe that layer drop affects MHA and PoM slightly differently, with the largest gains obtained on \textit{test-other} for MHA, and on \textit{test-clean} for PoM. % while changes on test-other are small. 
%Note that we trained all SummaryMixing models with layer drop.

\begin{table}[t] 
\centering
\caption{WER averaged by PoM version, with standard deviation.}
\label{tab:ablation_pom}
%\label{tab:ablation_training}
\resizebox{0.989\linewidth}{!}{
\begin{tabular}{l c r r r r}
\toprule
mixer & ldrop & clean & clean+LM & other & other+LM \\
\midrule
PoM base   & 1 & $8.76 \pm 0.28$ & \textbf{$5.51 \pm 0.15$} & \textbf{$19.87 \pm 0.51$} & \textbf{$14.02 \pm 0.32$} \\
PoM select & 1 & $8.74 \pm 0.20$ & $5.73 \pm 0.18$ & $20.56 \pm 0.60$ & $14.89 \pm 0.68$ \\
PoM 2ways  & 1 & $8.75 \pm 0.26$ & $5.65 \pm 0.11$ & $20.28 \pm 0.57$ & $14.60 \pm 0.31$ \\
PoM 3ways  & 1 & $8.77 \pm 0.47$ & $5.82 \pm 0.11$ & $20.86 \pm 0.76$ & $15.08 \pm 0.42$ \\
\bottomrule
\end{tabular}
}
\end{table}

\begin{table}[t]
\centering
\caption{WER averaged by mixer type, depending on the use of a language model at inference (LM) and the use of layerdrop during training (ldrop). (0: without, 1: with). Values in parentheses are absolute deltas vs. the same mixer’s baseline (LM=0, ldrop=0). Grey cells indicate the mixer-specific reference values.}
\label{tab:ablation_training}
\resizebox{0.989\linewidth}{!}{
\begin{tabular}{l c r r r r}
\toprule
mixer & ldrop & clean & clean+LM & other & other+LM \\
\midrule
PoM (all) & 0 & \refcell{8.92} & 5.71 ($-3.21$) & \refcell{19.86} & 14.12 ($-5.74$) \\
PoM (all) & 1 & 8.61 ($-0.31$) & 5.42 ($-3.50$) & 19.81 ($-0.05$) & 14.00 ($-5.86$) \\
\midrule
MHA & 0 & \refcell{8.39} & 5.30 ($-3.09$) & \refcell{19.01} & 13.25 ($-5.76$) \\
MHA & 1 & 8.20 ($-0.19$) & 5.05 ($-3.34$) & 18.19 ($-0.82$) & 12.53 ($-6.48$) \\
\bottomrule
\end{tabular}
}
\end{table}

\subsection{Perspectives}
To further balance expressivity and efficiency, we plan to investigate hybrid architectures with standard multi-head attention in the early layers and PoM in the upper layers. This is motivated by previous work~\cite{zhang2021usefulness}, which shows that upper-layer attention maps are mostly diagonal in ASR encoders and therefore need not model all token pairs exhaustively. 
We will also study architectural choices in PoM more finely, by varying, per layer, the polynomial order $k$, the expansion factor $D$, and the fraction of features participating in mixing, and by integrating PoM into different Transformer variants. 
Finally, we plan to benchmark PoM on additional downstream tasks (intent classification, emotion recognition, and speaker verification) and in streaming settings.

\section{Conclusion}
\label{sec:conclusion}
We introduced PoM, a novel token mixer with linear complexity in the number of input tokens. 
Replacing classic MHA with PoM in speech encoders and following the BEST-RQ pre-training method, we showed that PoM can be used as a drop-in replacement for attention while retaining sufficient expressivity for the complexity of spoken language.
PoM achieves WER rates close to MHA on the Librispeech-100h ASR task. PoM also surpasses SummaryMixing, its closest competitor of linear complexity, while achieving comparable runtime and memory use. 
In future work, we plan to further refine the PoM components and optimize their implementation for embedded devices.

% References 
% -------------------------------------------------------------------------
\bibliographystyle{IEEEbib}
\bibliography{refs}

\end{document}